\documentclass[twoside]{article}

\usepackage[accepted]{aistats2019}
\usepackage[round]{natbib}

\usepackage{graphicx}
\usepackage{amsmath,amssymb, verbatim}
\usepackage{subcaption}
\usepackage{algorithm, algorithmic}
\PassOptionsToPackage{hyphens}{url}\usepackage{hyperref}

\begin{document}

%

%

\twocolumn[

\aistatstitle{Probabilistic Semantic Inpainting with Pixel Constrained CNNs}

\aistatsauthor{ Emilien Dupont \And Suhas Suresha }

\aistatsaddress{ Schlumberger Software Technology Innovation Center } ]

\begin{abstract}
  Semantic inpainting is the task of inferring missing pixels in an image given surrounding pixels and high level image semantics. Most semantic inpainting algorithms are deterministic: given an image with missing regions, a single inpainted image is generated. However, there are often several plausible inpaintings for a given missing region. In this paper, we propose a method to perform probabilistic semantic inpainting by building a model, based on PixelCNNs, that learns a distribution of images conditioned on a subset of visible pixels. Experiments on the MNIST and CelebA datasets show that our method produces diverse and realistic inpaintings. 
\end{abstract}


\section{Introduction}

\begin{figure}[t]
\begin{center}
\includegraphics[width=0.8\linewidth, clip=true]{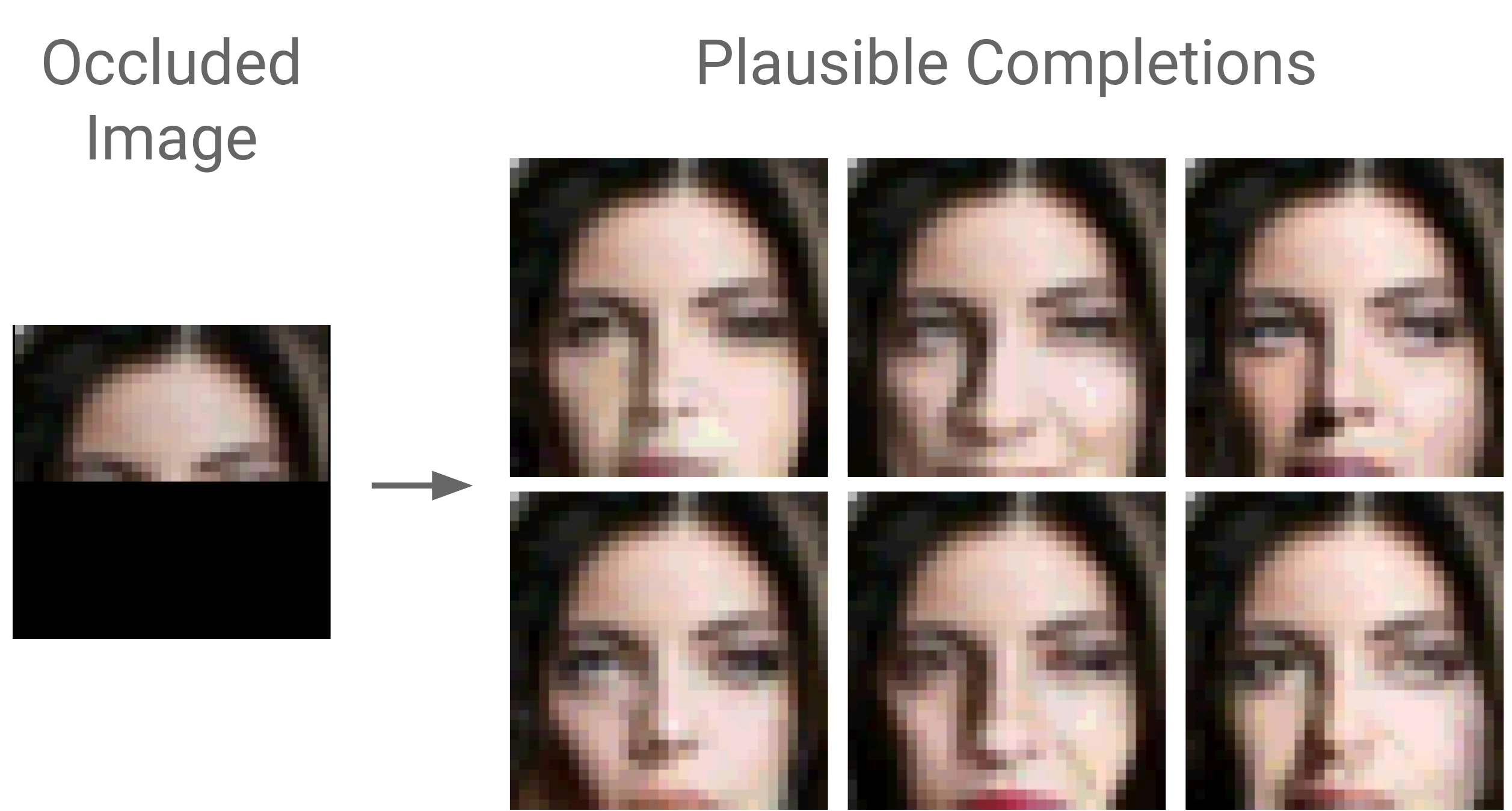}
\end{center}
\caption{Example of probabilistic inpainting.}
\label{intro-fig}
\end{figure}

Image inpainting algorithms find applications in many domains such as the restoration of damaged paintings and photographs \citep{bertalmio2000image}, the removal or replacement of objects in images \citep{liu2018image} or the generation of maps from sparse measurements \citep{dupont2018generating}. In these applications, a partially occluded image is passed as input to an algorithm which generates a complete image constrained by the visible pixels of the original image. As the missing or hidden regions of the image are unknown, there is an inherent uncertainty related to the inpainting of these images. For each occluded image, there are typically a large number of plausible inpaintings which both satisfy the constraints of the visible pixels and are realistic (see Fig. \ref{intro-fig}). As such, it is desirable to \textit{sample} image inpaintings as opposed to generating them deterministically. Even though recent algorithms have shown great progress in generating realistic inpaintings, most of these algorithms are deterministic \citep{liu2018image,yeh2016semantic, yu2018free}. 

In this paper, we propose a method to sample inpaintings from a distribution of images conditioned on the visible pixels. Specifically, we propose a model that simultaneously (a) generates realistic images, (b) matches pixel constraints and (c) exhibits high sample diversity. Our method, which we term Pixel Constrained CNN, is based on a modification of PixelCNNs \citep{van2016conditional,oord2016pixel} to allow for conditioning on arbitrary sets of pixels as opposed to only the ones above and to the left as in the original PixelCNN framework. 

Further, our model can estimate the likelihood of generated inpaintings. This allows us, for example, to rank the various generated inpaintings by their likelihood. To the best of our knowledge, this is the first method for efficiently estimating the likelihood of inpaintings of arbitrary missing pixel regions.

To validate our method, we perform experiments on the MNIST and CelebA datasets. Our results show that the model learns to generate realistic inpaintings while exhibiting high sample diversity. We also evaluate our method both qualitatively and quantitatively and show that it compares favourably with current state of the art methods. Finally, we measure, through a user survey, how well our likelihood estimates correlate with human perception of plausible inpaintings.



\section{Related Work}

Early approaches for image inpainting were mostly based on propagating the information available in the occluded image. Methods based on minimizing the total variation, for example, are able to fill small holes in an image \citep{shen2002mathematical,afonso2011augmented}. Other methods directly propagate information from visible pixels to fill in hidden pixel regions \citep{bertalmio2000image,ballester2001filling,telea2004image}. As these methods use only the information available in the image, they are unable to fill in large holes or holes where the color or texture have high variance. More importantly, these algorithms are also deterministic and so generate a single inpainting given an occluded image.


Other methods are based on finding patches in the occluded image or in other image datasets to infill the hidden regions \citep{efros2001image,kwatra2005texture}. This family of methods also includes PatchMatch \citep{barnes2009patchmatch} which has a random component. This randomness is however limited by the possible matches that can be found in the available datasets.


Learning based approaches have also been popular for inpainting tasks \citep{iizuka2017globally,yang2017high,song2017image,li2017context,yu2018generative}. Importantly, these methods often learn a prior over the image distribution and can take advantage of both this information and the pixel information available in the occluded image. \cite{liu2018image} for example achieve impressive results using partial convolutions, but these approaches are deterministic and the inpainting operation often corresponds to a forward pass of a neural network. Our method, in contrast, is able to generate several samples given a single inpainting task.


Several methods for image inpainting are also based on optimizing the latent variables of generative models. \cite{pathak2016context,yeh2016semantic} for example, train a Generative Adversarial Network (GAN) on unobstructed images \citep{goodfellow2014generative}. Using the trained GAN, these algorithms optimize the latent variables to match the visible pixels in the occluded image. These methods are pseudo random in the sense that different initializations of the latent variable can lead to different minima of the optimization problem that matches the generated image with the visible pixels. However, the resulting completions are typically not diverse \citep{bellemare2017cramer}. Further, since the final images are generated by the GAN, the lack of diversity of samples sometimes observed in GANs can also be limiting \citep{arjovsky2017wasserstein}. Our approach, in contrast, is based on PixelCNNs which typically exhibit high sample diversity \citep{dahl2017pixel}. 


Recently, \cite{xu2018controllable} proposed a model for controllable image inpainting. Their approach also uses an autoregressive model to generate image completions, but their focus is on controlling generated content as opposed to sampling diverse plausible completions. If we view image inpainting as regressing a function to match the visible pixels in an image, then Neural Processes \citep{garnelo2018neural, kim2019attentive} can also be considered as tools for image inpainting. We compare our approach with these methods in Section \ref{experiments}.

\section{Review of PixelCNNs}
\label{pixel-cnn-review}
PixelCNNs \citep{oord2016pixel,van2016conditional} are probabilistic generative models which aim to learn a distribution of images $p(\mathbf{x})$. These models are based on sampling each pixel in an image conditioned on all the previously sampled pixels. Specifically, letting $\mathbf{x}$ denote the set of pixels of an $n$ by $n$ image and numbering the pixels from 1 to $n^2$ row by row (in raster scan order), we can write $p(\mathbf{x})$ as:

\begin{equation}
\begin{split}
p(\mathbf{x}) & = p(x_1, x_2, ..., x_{n^2})\\
 & = \prod_{i=1}^{n^2} p(x_i | x_{i-1}, ..., x_1) \\
\end{split}
\end{equation}

We can then build a model for each pixel $i$, which takes as input the previous $i-1$ pixels and outputs the probability distribution $p(x_i | x_{i-1}, ..., x_1)$. We could, for example, build a CNN which takes as input the first $i-1$ pixels of an image and outputs the probability distribution over pixel intensities for the $i$th pixel. PixelCNNs use a hierarchy of masked convolutions to enforce this conditioning order, by masking pixels to the bottom and the right of each pixel, so that each pixel $i$ can only access information from pixels $i-1, i-2, ...$. The model is then trained by maximizing the log likelihood on real image data.

PixelCNNs are not only used to estimate $p(\mathbf{x})$ but also to generate samples from $p(\mathbf{x})$. To generate a sample, we first initialize all pixels of an image to zero (or any other number). After a forward pass of the image through the network, the output at pixel 1 is the distribution $p(x_1)$. The value of the first pixel of the image can then be sampled from $p(x_1)$.
After setting pixel 1 to the sampled value, we pass the image through the network again to sample from $p(x_2|x_1)$. We then set pixel 2 to the sampled value and repeat this procedure until all pixels have been sampled.

However, PixelCNNs can only generate images in raster scan order. For example, if pixel 3 is known, then we cannot sample $p(x_2 | x_1, x_3)$ since this does not match the sampling order imposed by the masking. In image inpainting, an arbitrary set of pixels is fixed and known, so we would like to be able to sample from distributions conditioned on any subset of pixels. A trivial way to enforce this conditioning is to modify the PixelCNN architecture to take in the visible pixels as a conditioning vector (see Conditional PixelCNNs for more details \citep{van2016conditional}). However, our initial experiments showed that the conditioning is largely ignored and the model tends to generate images which do not match the conditioning pixels. Similar problems have been observed when using PixelCNNs for super resolution \citep{dahl2017pixel}.

\section{Pixel Constrained CNN}
In this section we introduce Pixel Constrained CNN, a probabilistic generative model that can generate image samples conditioned on arbitrary subsets of pixels. Specifically, given a set of known constrained pixel values $\mathbf{c}$ (e.g. $\mathbf{c}=\{x_{17}, x_{52}, x_{134}\}$) we would like to model and sample from $ p(\mathbf{x} | \mathbf{c})$, i.e. we would like to sample all the pixels $\mathbf{x}$ in an image, given the visible pixels $\mathbf{c}$. We factorize $p(\mathbf{x} | \mathbf{c})$ as

\begin{equation}
p(\mathbf{x}| \mathbf{c}) = \prod_{i : x_{i} \notin \mathbf{c}} p(x_i | x_{i-1}, ..., x_1, \mathbf{c})
\end{equation}

where the product is over all the missing pixels in the image. As noted in section \ref{pixel-cnn-review}, PixelCNNs enforce this factorization by hiding pixels with a hierarchy of masked convolutions. In the constrained pixel case, we would like to hide pixels in the same order \textit{except} for the known pixels $\mathbf{c}$ which should be visible to all output pixel distributions. Therefore, building the constrained model amounts to using the same factorization as the original PixelCNN, but modifying the masking to make the constrained pixels visible to all pixels. This can be achieved by building a model composed of two subnetworks, a prior network and a conditioning network.

The prior network is a PixelCNN, which takes as input the full image and outputs logits which encode information from pixels $i-1, i-2, ...$ for each pixel $i$. The conditioning network is a CNN with regular (non masked) convolutions which takes as input the masked image, containing only the visible pixels $\mathbf{c}$, and outputs logits which encode the information in the visible pixels. Since the conditioning network does not use masked convolutions, each pixel in the logit output will have access to every visible pixel in the input (assuming the network is deep enough for the receptive field to cover the entire image).

\begin{figure}[t]
\begin{center}
\includegraphics[width=0.8\linewidth, clip=true]{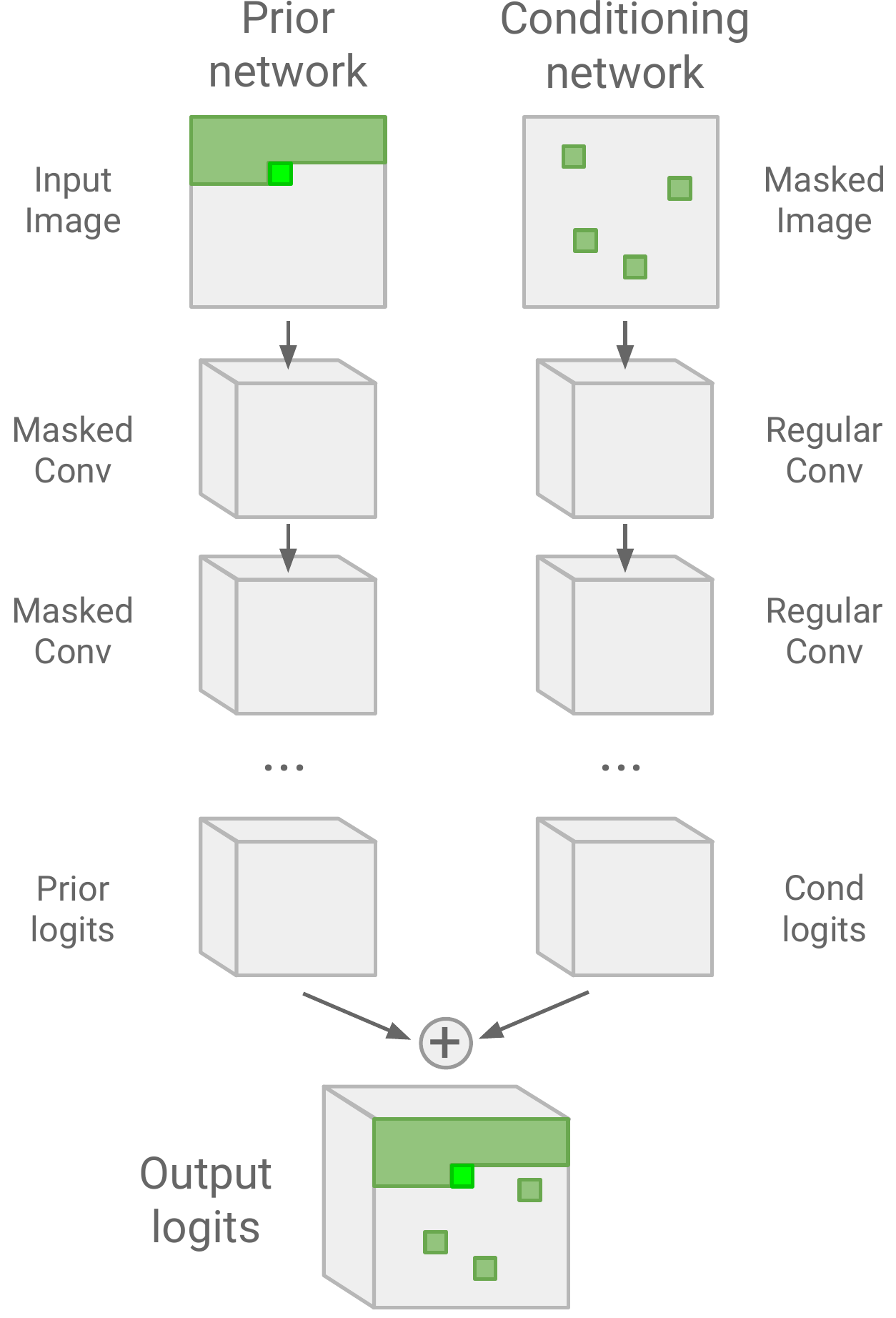}
\end{center}
\caption{Pixel Constrained CNN architecture. The complete input image is passed through the prior network and the masked image is passed through the conditioning network. The light green pixel corresponds to the $ith$ pixel which can access information from all the pixels in dark green, i.e $x_{i-1}, x_{i-2}, ..., x_0$ and $\mathbf{c}$.}
\label{architecture}
\end{figure}

Finally, the prior logits and conditional logits are added to output the final logits. The softmax of these logits models the probability distribution $p(x_i | x_{i-1}, ..., x_1, \mathbf{c})$ for each pixel $i$. This approach is illustrated in Fig. \ref{architecture}. Note that a similar approach has been used in the context of super resolution, where the conditioning network takes in a low resolution image instead of a masked image \citep{dahl2017pixel}.

\begin{figure}[t]
\begin{center}
\includegraphics[width=1.0\linewidth, clip=true]{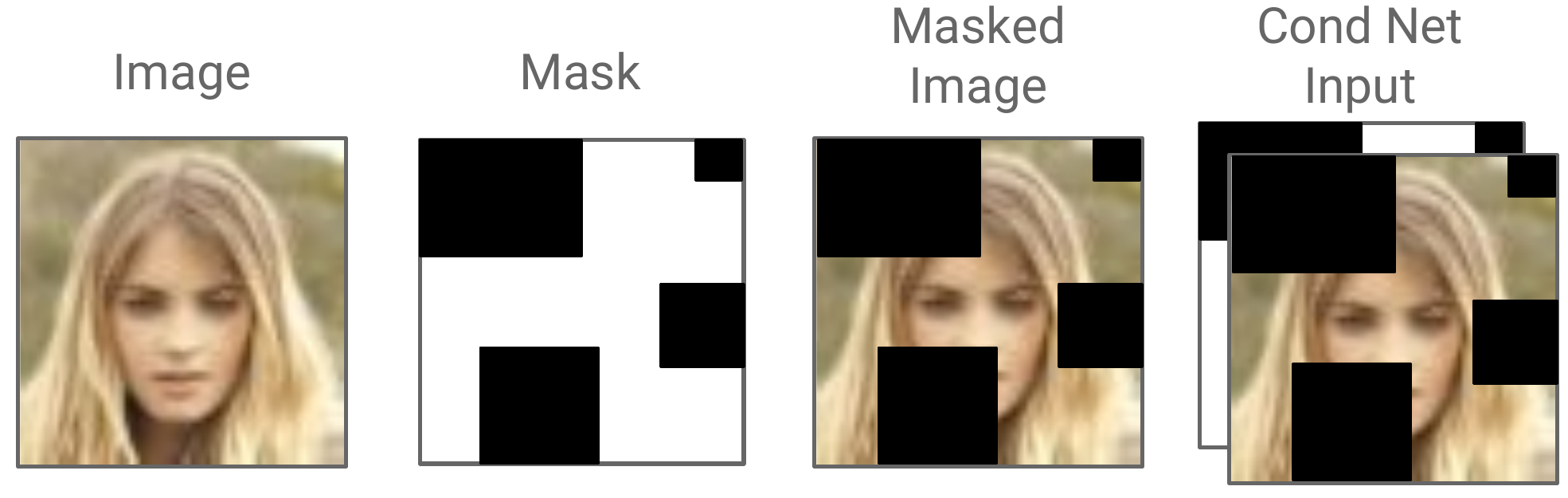}
\end{center}
\caption{Conditioning network input. The input is a concatenation of the mask and the masked image in order to differentiate between hidden pixels and visible pixels which have a value of zero. This input represents the set of constrained pixels $\mathbf{c}$.}
\label{cond-input}
\end{figure}

\begin{figure}[h]
\begin{center}
\includegraphics[width=1.0\linewidth, clip=true]{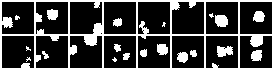}
\end{center}
\caption{Examples of 32 by 32 masks generated by our algorithm. These correspond to samples from the distribution of masks $p(M)$.}
\label{mask-examples}
\end{figure}

\subsection{Model Inputs}

During training, the prior network takes as input a fully visible image while the conditioning network takes as input a masked version of the same image, representing the constrained pixels $\mathbf{c}$. More precisely, the constrained pixels $\mathbf{c}$ can be thought of as a function $\mathbf{c}(\mathbf{x}, M)$ of the image $\mathbf{x}$ and a mask $M$. The mask $M \in \{0, 1 \}^{n \times n}$ is a binary matrix, where 1 represents a visible pixel and 0 a hidden pixel. The input image $\mathbf{x} \in \mathbb{R}^{n \times n \times c}$ (where $c$ is the number of color channels) is masked by elementwise multiplication with $M$. To differentiate between masked pixels and pixels which are visible but have a value of 0, we append $M$ to the masked image, so the final input to the conditioning network is in $\mathbb{R}^{n \times n \times (c + 1)}$. This is illustrated in Fig. \ref{cond-input}. The approach is similar to the one used by \cite{zhang2016colorful} for deep colorization.

\subsection{Likelihood Maximization}

We train the model by maximizing $p(\mathbf{x} | \mathbf{c})$ on a dataset of images. Ideally, the trained model should work for any set of constrained pixels $\mathbf{c}$ or, equivalently, for any mask. 
To achieve this, we define a distribution of masks $p(M)$ and maximize the log likelihood of our model over both the masks and the data

\begin{equation}
\label{theory-loss}
\max \mathbb{E}_{\mathbf{x} \sim p(\mathbf{x}), M \sim p(M)}[\log p(\mathbf{x}| \mathbf{c}(\mathbf{x}, M)] 
\end{equation}

When optimizing this loss in practice, we found that the conditional logits (which model the information in $\mathbf{c}$) were often partially ignored by the model. We hypothesize that this is because there is a stronger correlation between a pixel and its neighbors (which is what the prior network models) than between a pixel and the visible pixels in the image (which is what the conditioning network models). To encourage the model to use the conditional logits, we add a weighted term to the loss. Denoting by $p_\text{cond}(\mathbf{x} | \mathbf{c})$ the softmax of the conditional logits, the total loss is

\begin{equation}
\max \mathbb{E}_{\mathbf{x} \sim p(\mathbf{x}), M \sim p(M)}[\log p(\mathbf{x}| \mathbf{c}) + \alpha \log p_\text{cond}(\mathbf{x}| \mathbf{c})] 
\end{equation}

where $\alpha$ is a positive constant and the dependence of $\mathbf{c}$ on $\mathbf{x}$ and $M$ has been omitted for clarity. This loss encourages the model to encode more information into the conditional logits and we observe that this improves performance in practice.

\subsection{Random Masks}
In order to evaluate the loss and train the model, we need to define the distribution of masks $p(M)$. There are several ways this can be done. For example, if it is known a priori that we are only interested in completing images which have part of their right side occluded, we can train on masks of varying width covering the right side of the image. While this is application dependent, we would like to build models that are as general as possible and can work on a wide variety of masks. Specifically, we would like our model to perform well even when missing areas are irregular and disconnected. To this end, we build an algorithm that generates irregular masks of random blobs. The algorithm randomly samples blob centers and then iteratively and randomly expands each blob. The algorithm is described in detail in the appendix and examples of the generated masks are shown in Fig. \ref{mask-examples}. While the generated masks are all irregular we find that they generalize well to completing any occlusion in unseen images, including regular occlusions.

\subsection{Sampling}
\label{sampling-likelihood-section}
Given a trained model and an image with a subset of visible pixels $\mathbf{c}$, we would like to generate samples from the distribution $p(\mathbf{x} | \mathbf{c})$. To generate these, we first pass the occluded image and the mask through the conditioning network to calculate the conditional logits. When then pass a blank image through the prior network to generate the prior logits for the first pixel. If the first pixel is part of the visible pixels $\mathbf{c}$, we simply set $x_1$ to the value given in $\mathbf{c}$, otherwise we sample $x_1 \sim p(x_1 | \mathbf{c})$ and set the value of the first pixel to $x_1$. We then pass the updated image through the prior network again to generate the conditional probability distribution for the second pixel and continue sampling in this way until the image is complete. Since we know the probability distribution at every pixel, we can also calculate the likelihood of the generated sample by taking the product of the probabilities of each sampled pixel.

\section{Experiments}
\label{experiments}

We test our model on the binarized MNIST\footnote{The images are binarized by setting any pixel intensity greater than 0.5 to 1 and others to 0.} and CelebA datasets \citep{liu2015faceattributes}. As training the model is computationally intensive, we crop and resize the CelebA images to 32 by 32 pixels and quantize the colors to 5 bits (i.e. 32 colors). For both MNIST and CelebA, we use a Gated PixelCNN \citep{van2016conditional} for the prior network and a residual network \citep{he2016deep} for the conditioning network. Full descriptions of the network architectures are given in the supplementary material.

Since generating masks at every iteration is expensive, we generate a dataset of 50k masks prior to training and randomly sample these during training. The full list of training details can be found in the supplementary material. The code to reproduce all experiments and results in this paper (including weights of trained models) is available at \url{https://github.com/Schlumberger/pixel-constrained-cnn-pytorch}.

\begin{figure}[t]
\begin{center}
\includegraphics[width=1.0\linewidth]{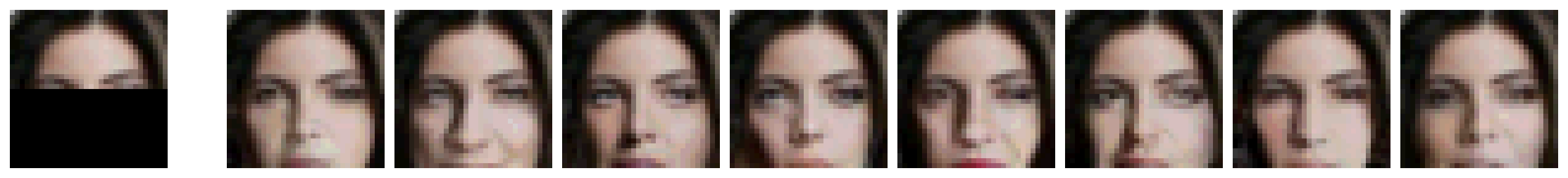}
\includegraphics[width=1.0\linewidth]{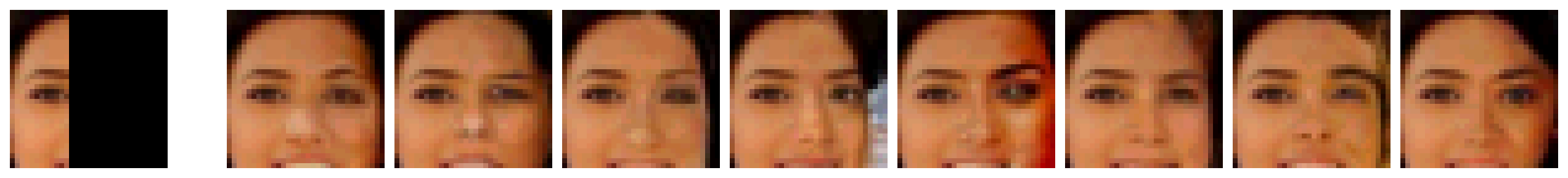}
\includegraphics[width=1.0\linewidth]{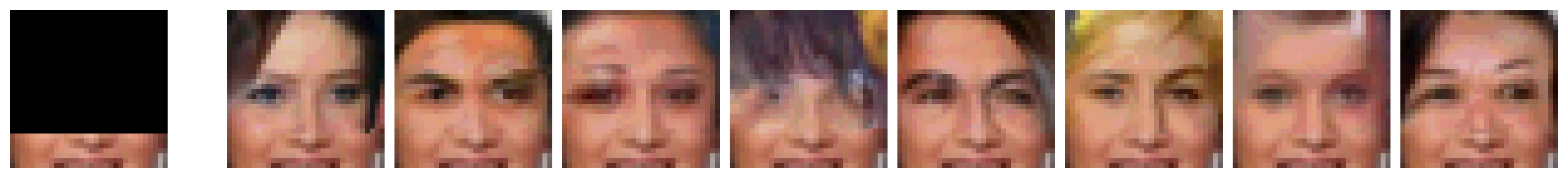}
\includegraphics[width=1.0\linewidth]{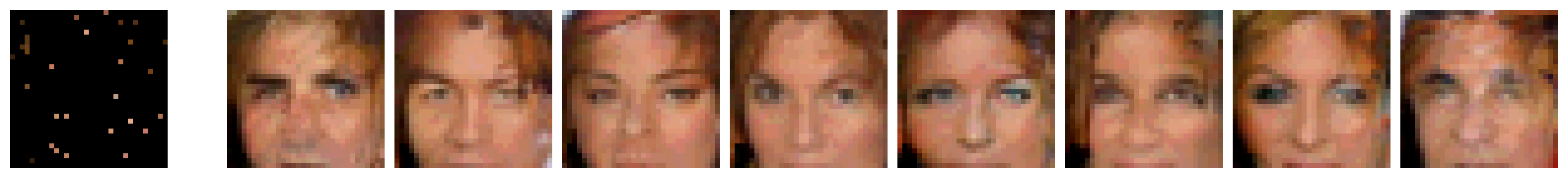}
\includegraphics[width=1.0\linewidth]{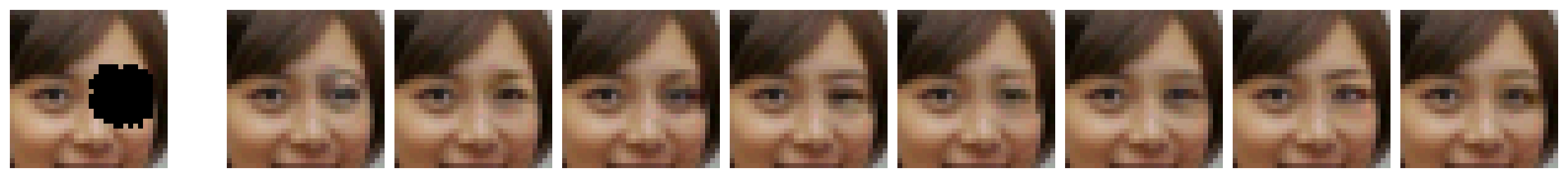}
\end{center}
\caption{Inpainting results on CelebA. The occluded images are shown on the left and various completions sampled from the model are shown on the right. As can be seen, the samples are diverse and mostly realistic. }
\label{inpainting-results}
\end{figure}

\begin{figure}[t]
\begin{center}
\includegraphics[width=1.0\linewidth]{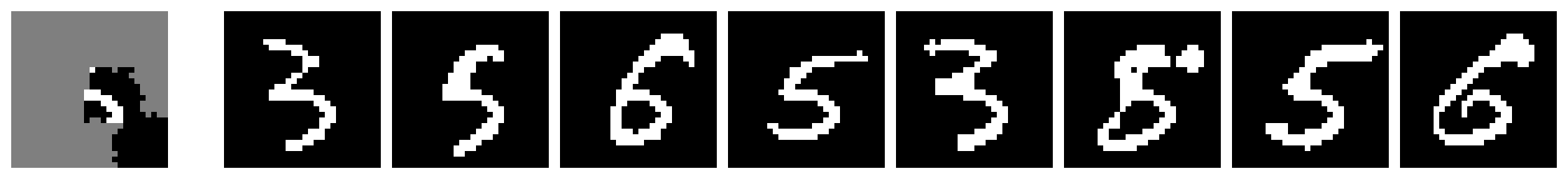}
\includegraphics[width=1.0\linewidth]{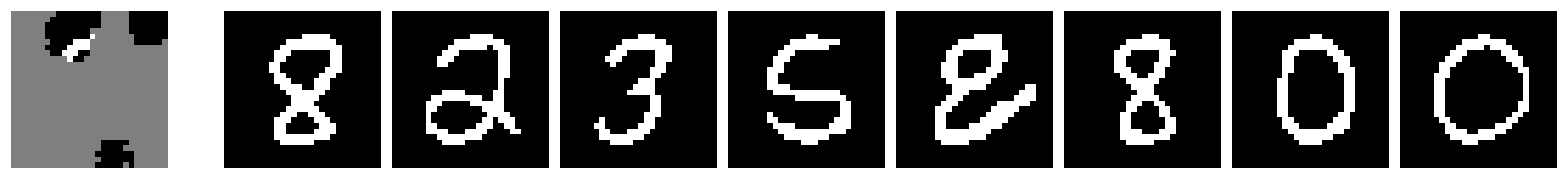}
\includegraphics[width=1.0\linewidth]{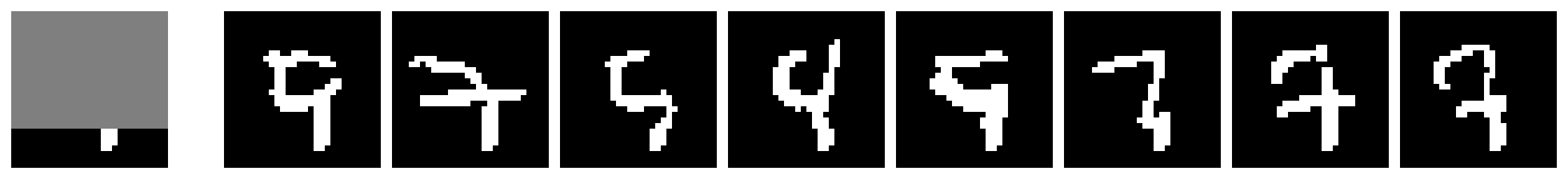}
\includegraphics[width=1.0\linewidth]{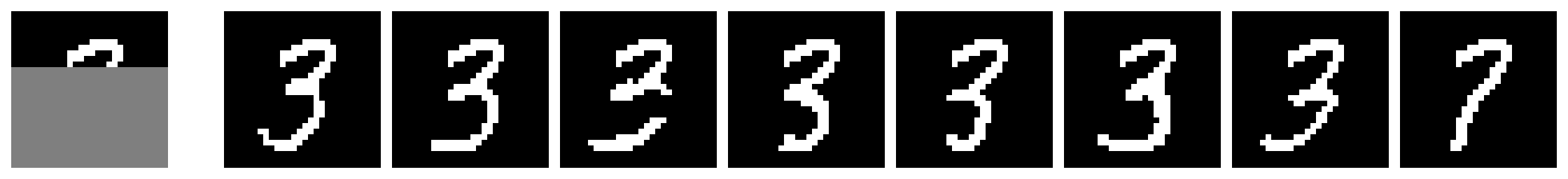}
\includegraphics[width=1.0\linewidth]{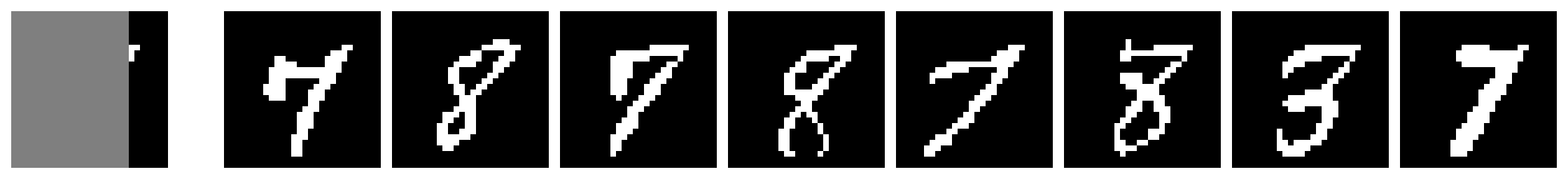}
\end{center}
\caption{Inpainting results on MNIST.}
\label{inpainting-results-mnist}
\end{figure}

\begin{figure}[t]
\begin{center}
\includegraphics[width=0.49\linewidth]{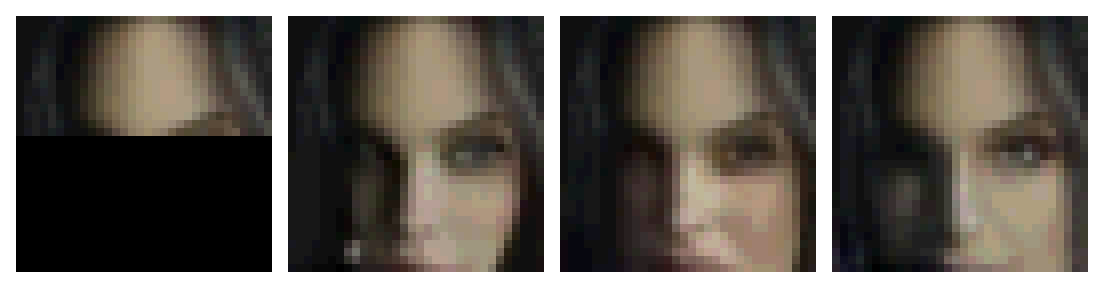}
\includegraphics[width=0.49\linewidth]{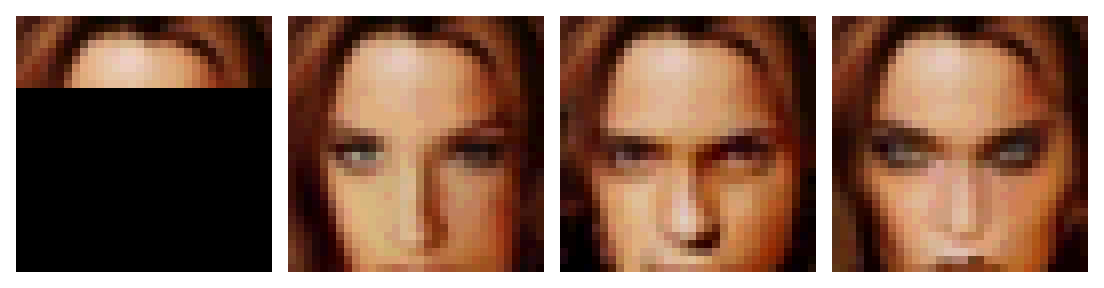}
\includegraphics[width=0.49\linewidth]{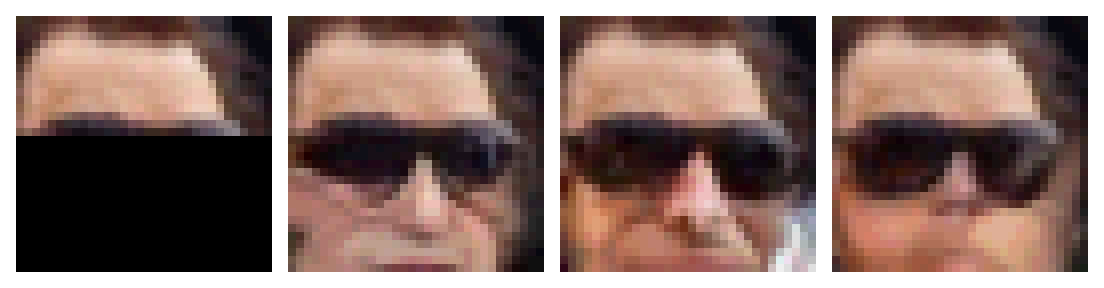}
\includegraphics[width=0.49\linewidth]{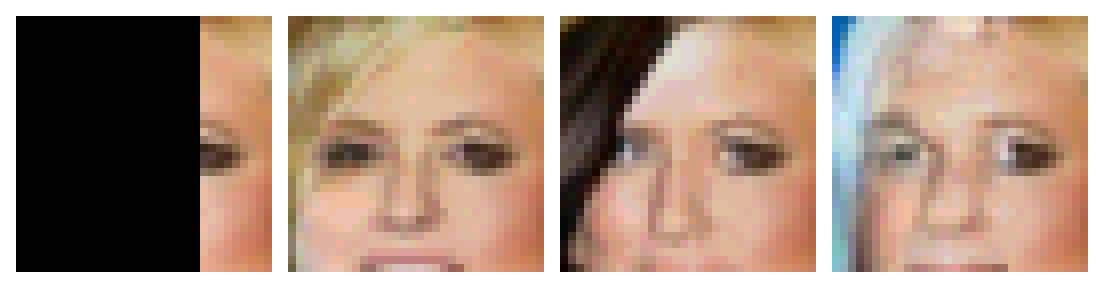}
\includegraphics[width=0.49\linewidth]{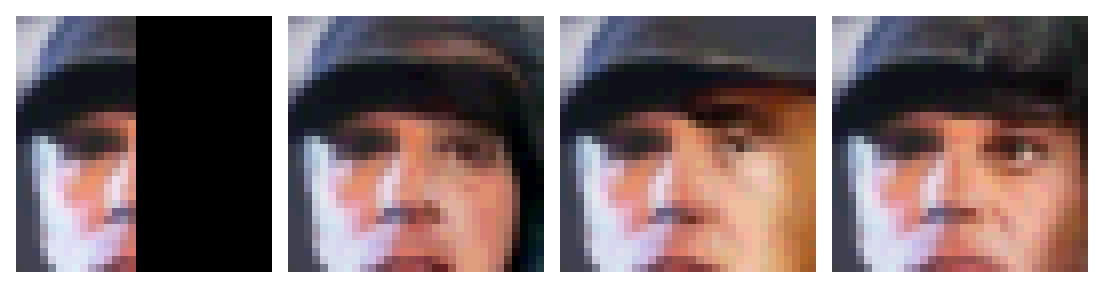}
\includegraphics[width=0.49\linewidth]{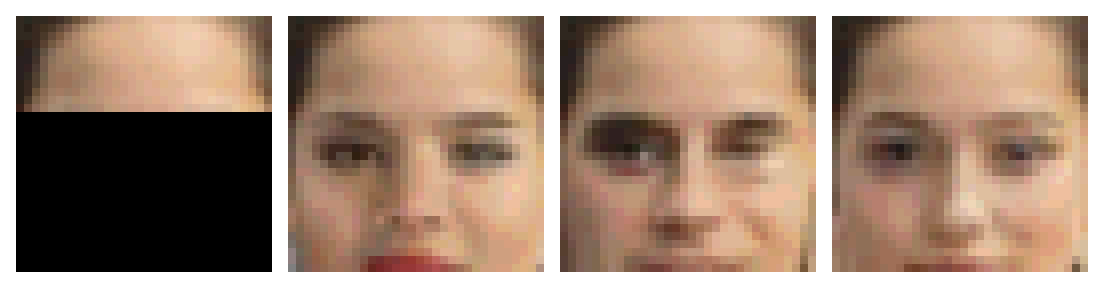}
\end{center}
\caption{Inpainting examples. These examples showcase various interesting aspects of the model. For example, the model can inpaint various types of glasses, hats and shadows. It also produces diverse eye color, hair color and gender for a given occluded image.}
\label{interesting-fig}
\end{figure}

\begin{figure}[t]
\begin{center}
\includegraphics[width=0.9\linewidth]{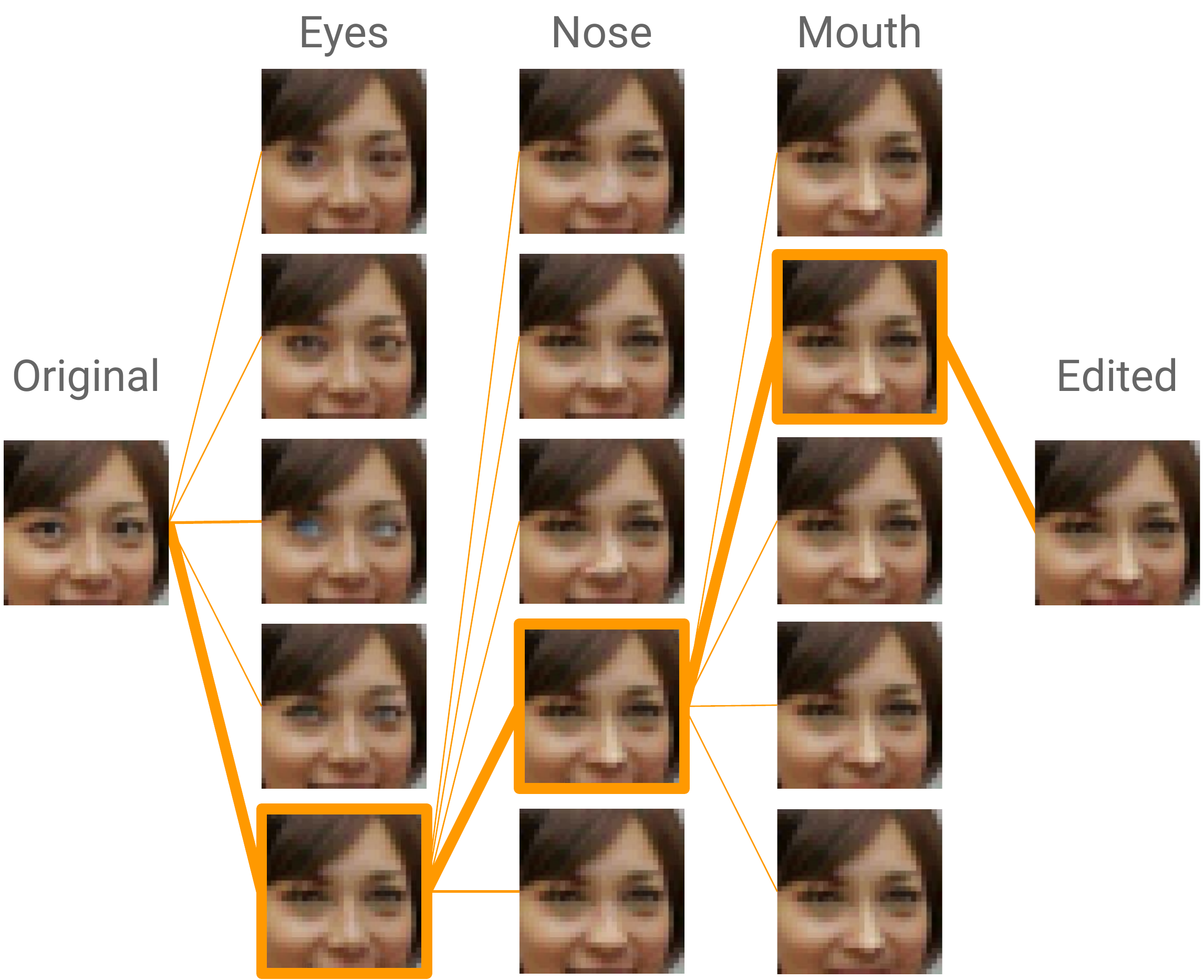}
\end{center}
\caption{Probabilistic inpainting for image editing. We start with the image on the left and sample new eyes, a new nose and a new mouth to edit the image.}
\label{image-editing}
\end{figure}

\begin{figure*}[h]
\begin{center}
\includegraphics[width=0.9\textwidth]{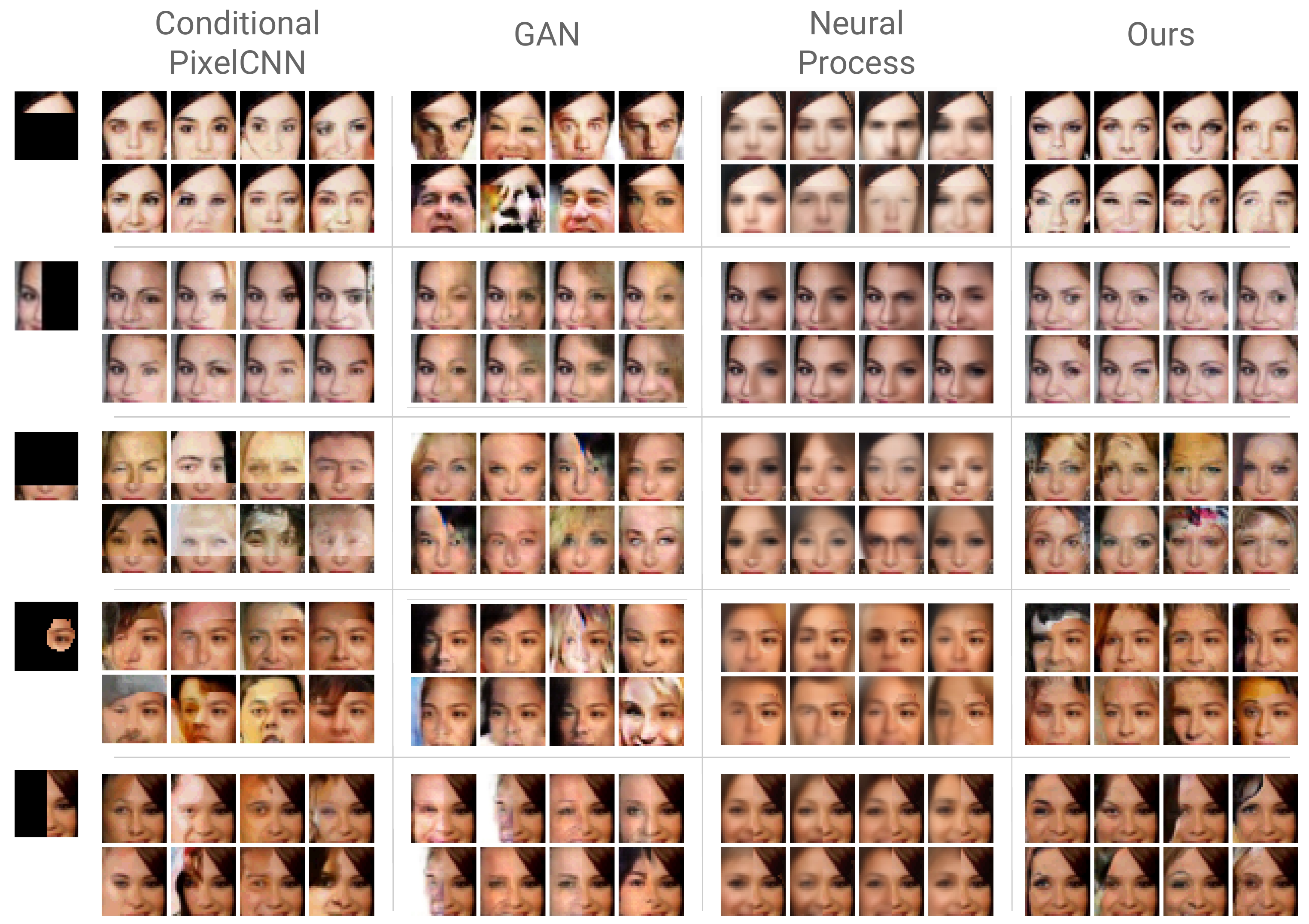}
\end{center}
\caption{Comparisons of methods on various inpainting tasks. The occluded images are shown on the left and various completions sampled from the models are shown on the right in each column.}
\label{comparisons-fig}
\end{figure*}

\subsection{Inpaintings}

We test our models on images and masks that were not seen during training. Examples of inpaintings are shown in Fig. \ref{inpainting-results} and \ref{inpainting-results-mnist}. As can be seen, the generated samples are realistic and, importantly, diverse. For example, even when the source image for the masked pixels is of a male face, the model plausibly generates a variety of both male and female face completions, each with varying hair, eye color, skin tone and so on. Importantly, the inpaintings also preserve the structure in the image: facial symmetry is preserved when either side of a face is occluded, eye color is preserved when an eye is occluded and so on (interestingly, the model occasionally samples a blue eye even when the other eye is dark). 

For MNIST, we observe similar results. The model generates a variety of digits, all of which naturally match the conditioning. Interestingly, even when the digit used to generate the visible pixels is a seven, the model is able to generate many other digit completions which plausibly match the constrained pixels.

Further examples of inpaintings and an application of probabilistic inpainting to image editing are shown in Fig. \ref{interesting-fig} and \ref{image-editing}.

\subsection{Qualitative Evaluation}

In order to evaluate the performance of our model, we run qualitative comparisons with Conditional PixelCNNs \citep{van2016conditional}, GAN inpainting \citep{yeh2016semantic} and Neural Processes \citep{garnelo2018neural}, each of which can be used to perform inpainting with some randomness. We compare the frameworks on different inpainting tasks by occluding 5 images with various masks (top, bottom, right, left and blob) and generating 8 inpaintings for each image and mask combination, allowing us to check for both diversity and plausibility. Results are shown in Fig. \ref{comparisons-fig}. As can be seen, Conditional PixelCNNs tend to produce images that do not match the visible pixels, indicating that the model largely ignores the conditional vector. This effect is particularly severe when generating inpaintings based on the bottom pixels, as the model can then only use the conditional vector to model this information (since the pixel ordering for PixelCNNs is from top to bottom). GAN inpainting is better at matching visible pixels, but still struggles on most tasks and has lower diversity. Neural Processes generate samples with some diversity, but these tend to be blurry and so fail to match the visible pixels. In contrast, our model produces plausible and diverse samples which match the pixels for all inpainting tasks. While it can occasionally produce unrealistic samples, our comparisons clearly show the value of our method for probabilistic inpainting.

\subsection{Quantitative Evaluation}

As noted in \cite{yu2018generative}, it is difficult to quantitatively measure the quality of image inpaintings. Currently, standard methods include simple metrics such as $\ell_1$, $\ell_2$ and pSNR between the inpainted image and the ground truth. As our model is probabilistic, it is even more difficult to use these metrics, since a plausible inpainting may be far from the ground truth image. Nonetheless, we still perform quantitative comparisons as follows: we sample 100 images randomly from CelebA and sample a mask randomly for each image. We then generate 8 inpaintings for each image and mask pair and measure the differences between the ground truth image and the generated image. Results are included in the table below. As can be seen, our model performs on par with other models on these metrics (although it is likely to perform worse than deterministic models since diversity is not measured). While measuring performance on these metrics is a good sanity check, we believe that designing a good metric for measuring inpainting quality is still an open problem.

\begin{center}
\begin{tabular}{ p{3em} | p{3em}| p{3em} | p{8em} }
 & GAN & NP & Ours\footnotemark \\
\hline
$\ell_1$ & 31.4\% & 19.9\% & 19.8\% (15.5\%) \\ 
\hline
$\ell_2$ & 40.6\% & 28.2\% & 29.0\% (22.4\%) \\ 
\hline
pSNR & 14.5dB & 17.4dB & 17.4dB (19.5dB) \\ 
\end{tabular}
\end{center}
\footnotetext{The results in brackets are for the best sampled inpainting for each image. Other results are averaged over all inpaintings for each image.}

\subsection{Inpainting Likelihood}

As noted in section \ref{sampling-likelihood-section}, our method allows us to calculate the likelihood of inpaintings. To the best of our knowledge, this is the first method for semantic inpainting of arbitrary occlusions which also estimates the likelihood of the inpaintings. The ability to estimate inpainting likelihoods could be useful for applications where the inpainted image is used for downstream tasks which require some uncertainty quantification \citep{dupont2018generating}. Ideally, inpaintings with low likelihood should look less realistic, while inpaintings with high likelihood should look more plausible. In this section, we perform various tests, including a human survey, to verify to which extent this claim is true. 

Firstly, we can verify that the model assigns high likelihood to the ground truth images. To test this, we generate 8 inpaintings for a 100 images with arbitrary occlusions. We then compare the likelihood assigned to the ground truth (unoccluded) image with the likelihood assigned to the inpaintings and rank them from highest to lowest, i.e. the best rank is 1 and the worst rank is 9. For MNIST, the ground truth image is ranked as number 1.62 on average. For CelebA, this number is 1.11. These results confirm that the model assigns high likelihood to the ground truth images, as expected.

\begin{figure}[t]
\centering
\begin{subfigure}[t]{0.9\linewidth}
\includegraphics[width=1.0\linewidth]{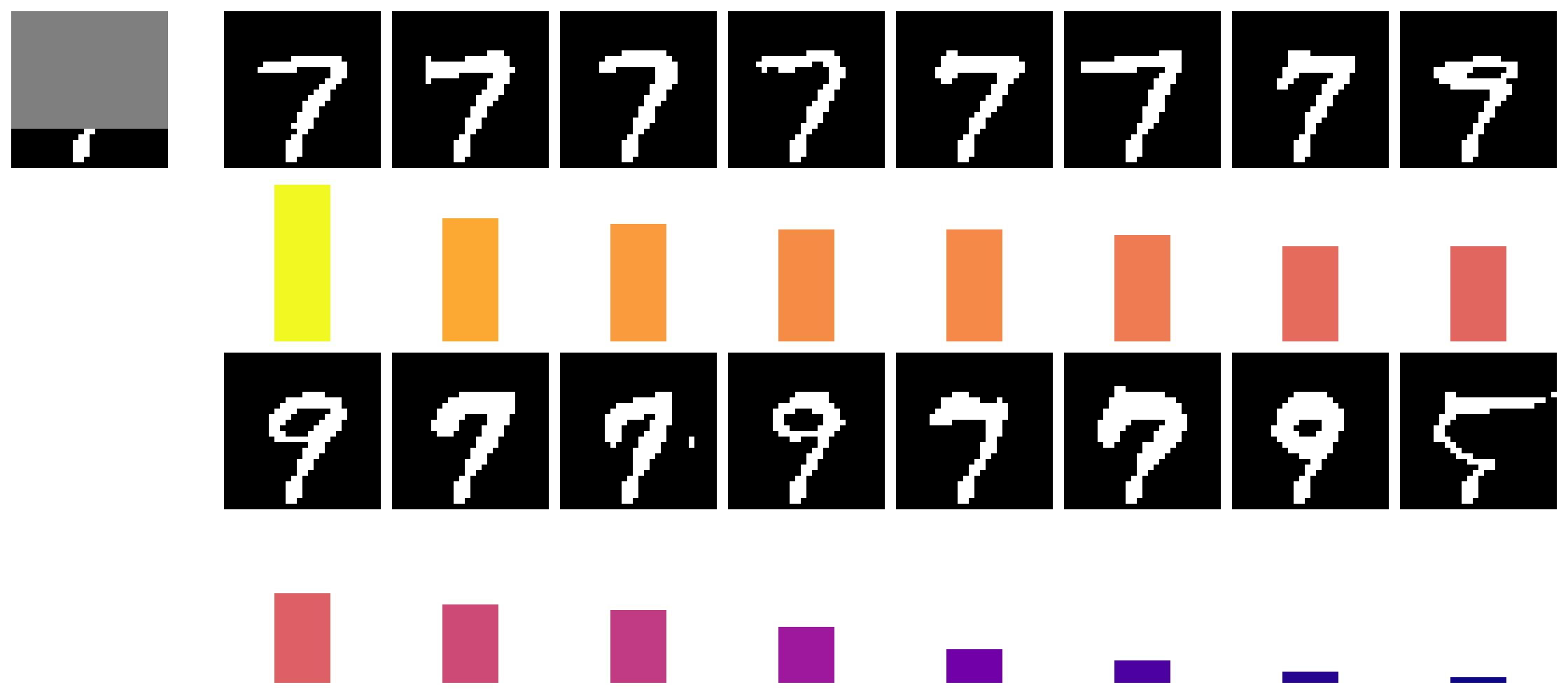}
\end{subfigure}
\begin{subfigure}[t]{0.9\linewidth}
\includegraphics[width=1.0\linewidth]{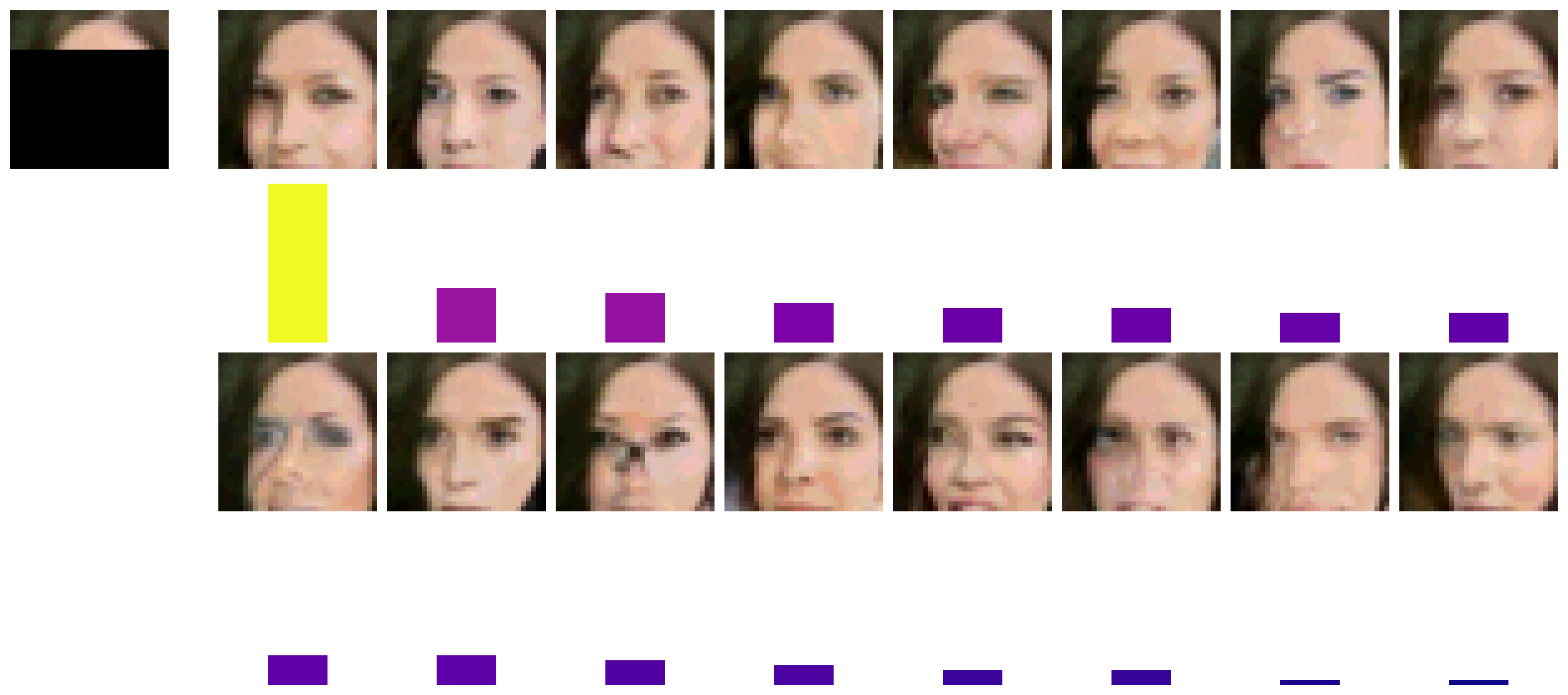}
\end{subfigure}
\caption{Inpaintings sorted by their likelihood. The occluded images are shown on the left and various completions sampled from the model are shown on the right. The size and color of the bar under each image is proportional to the likelihood of the sample.}
\label{inpainting-likelihood}
\end{figure}

Secondly, we can visualize how inpaintings with different likelihoods compare. Fig. \ref{inpainting-likelihood} shows a set of sampled inpaintings ranked by their likelihood. Even though the concept of how plausible an inpainting is can be subjective, it appears, at least for MNIST, that samples with high likelihood tend to look more plausible while low likelihood samples tend to look less realistic. For CelebA, this is not quite as clear. In order to test this more thoroughly, we conducted a user study.

\subsubsection{Human survey}

We performed a user study to check if our likelihood estimates correlate well with human perception of plausible inpaintings. The test works as follows: we occlude parts of randomly selected images and generate 8 inpaintings for each of them. For each image, we select the sample with highest and lowest likelihood and show this pair of inpaintings to the user (in random order). The user is then asked to decide which image they feel is most plausible. As the inpaintings can be quite similar, the user also has an option for choosing neither. If the likelihood estimate is good, the user should pick the image with the highest likelihood more frequently. The exact details of the experimental setup and examples of the user interface can be found in the supplementary material.

We collected responses from 43 users for MNIST and 42 users for CelebA resulting in a total of 1065 image pairs. For each dataset we then calculated the frequency with which users agree or disagree with the model and the frequency with which they replied "Don't know". The results are shown in the table below.

\begin{center}
\begin{tabular}{ p{4em} | p{4em}| p{4em} | p{5em} } \label{human-table}
 & Agree & Disagree & Don't know \\
\hline
MNIST & \textbf{49.7\%} & 25.5\% & 24.8\%\\ 
\hline
CelebA & 28.1\% & \textbf{56.6\%} & 15.3\% \\ 
\end{tabular}
\end{center}

For MNIST, the users agree with the model likelihood the majority of the time, suggesting our likelihood estimate correlates well with human perception. However, for CelebA the opposite is true. The users in our survey found that the majority of images the model assigned high likelihood to were less plausible. This suggests that the likelihood estimate for the CelebA model is flawed. We hypothesize that this is because the CelebA model has not completely converged or does not have enough capacity to accurately model the conditional distributions of faces. It is likely that training larger models could improve results but we leave this to future work. While estimating the likelihood or plausibility of inpaintings remains a challenge, we hope that this is a first step towards solving this problem.

\subsection{Pixel Probabilities}

\begin{figure}[t]
\begin{center}
\includegraphics[width=0.6\linewidth, clip=true]{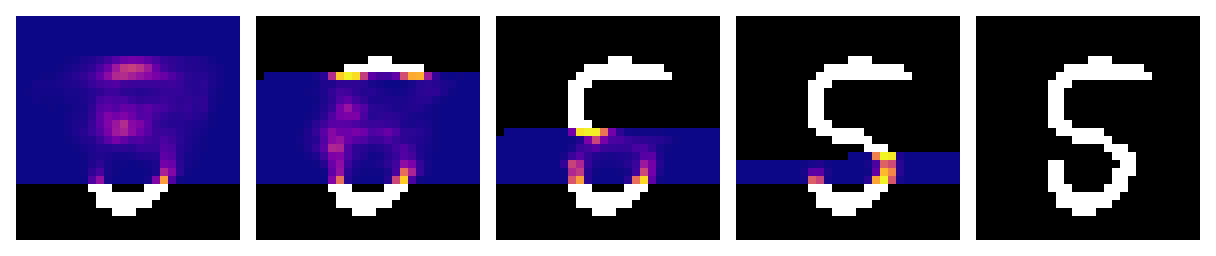}
\includegraphics[width=0.6\linewidth, clip=true]{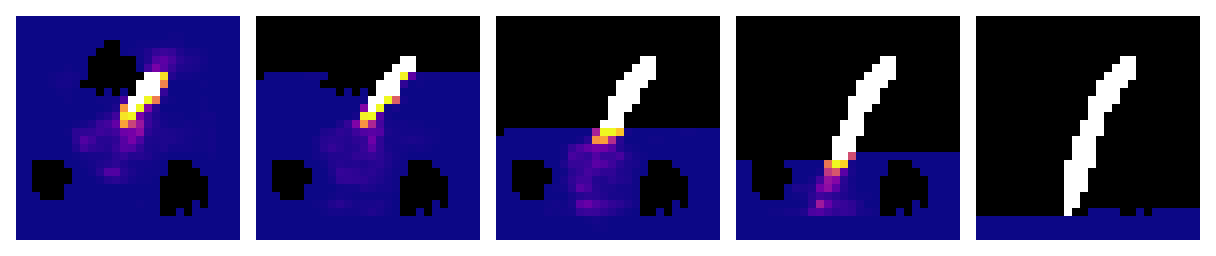}
\includegraphics[width=0.6\linewidth, clip=true]{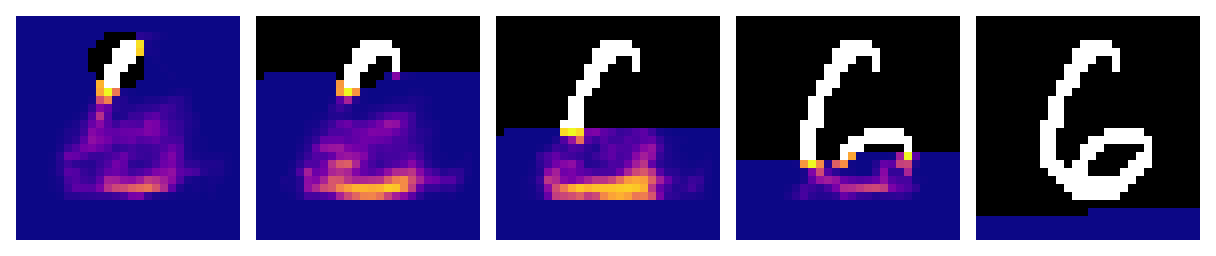}
\includegraphics[width=0.6\linewidth, clip=true]{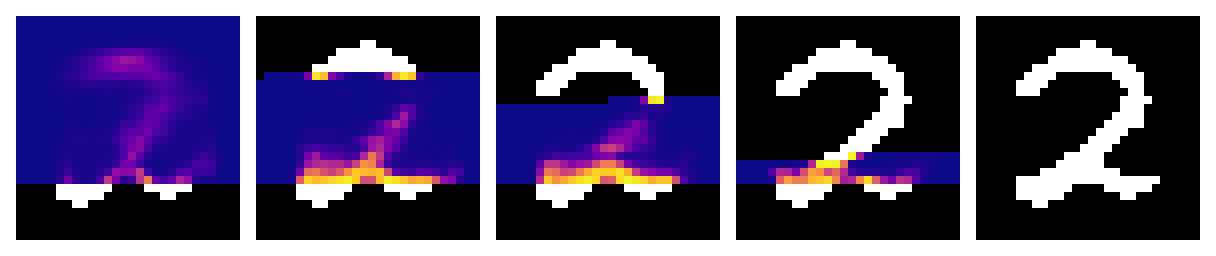}
\includegraphics[width=0.6\linewidth, clip=true]{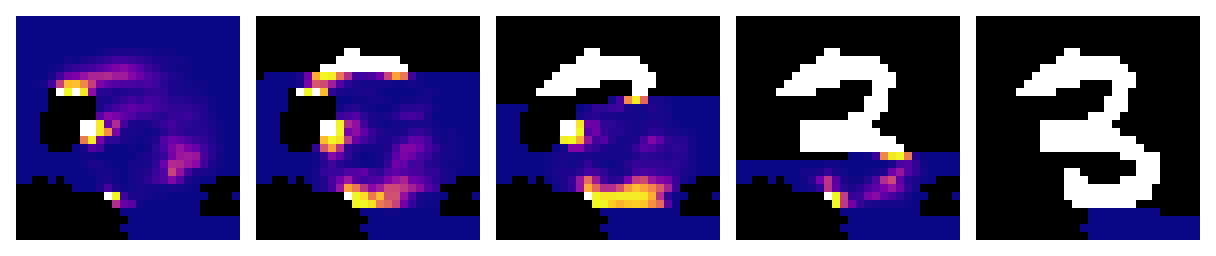}
\end{center}
\caption{Pixel probability progression as pixels are sampled (figure best viewed in color). The color of each hidden pixel is proportional to the probability of that pixel being 1 (bright colors correspond to high probabilities while dark colors correspond to low probabilities).}
\label{prob-progression}
\end{figure}

As our model estimates the probability for each pixel $p(x_i | x_{i-1}, ..., x_1, \mathbf{c})$, we can also visualize how the pixel probabilities are affected by various occlusions. Since the MNIST images are binary, we can plot the probability of a pixel intensity being 1 for all pixels in the image, given the visible pixels. Similarly, we can observe how these probabilities change as more pixels are sampled. This is shown in Fig. \ref{prob-progression}. As can be seen, the conditional pixels bias the model towards generating digits which are plausible given the occlusion. As more pixels are generated, the probabilities become sharper as the model becomes more certain of which digit it will generate. For example, in the first row, the pixel probabilities suggest that both a 3, 5 or an 8 are plausible completions. As more pixels are sampled it becomes clear that a 5 is the only plausible completion and the pixel probabilities get updated accordingly.

\section{Scope and Limitations}

While our approach can generate a diverse set of plausible image completions and estimate their likelihood, it also comes with some drawbacks and limitations. 

\begin{figure}[t]
\begin{center}
\includegraphics[width=0.8\linewidth, clip=true]{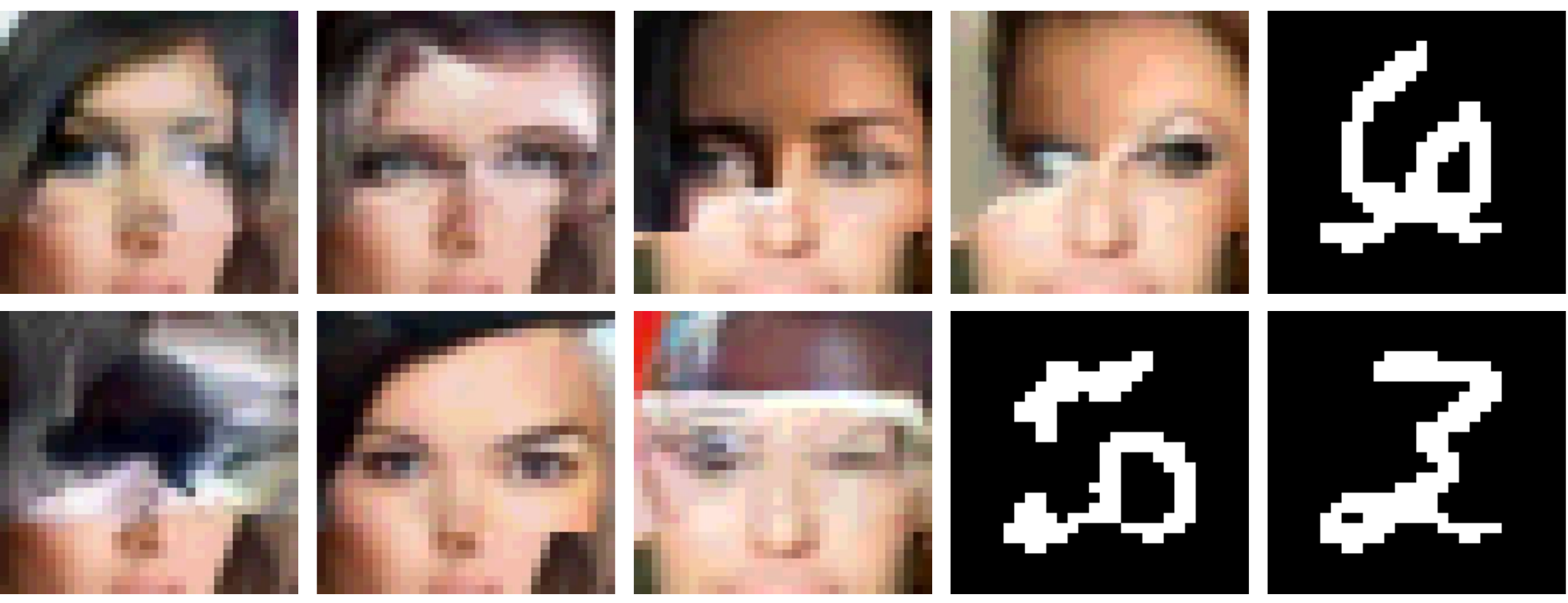}
\end{center}
\caption{Failure examples. The samples are either unrealistic or do not match the constrained pixels, creating undesirable edge effects.}
\label{failure-examples}
\end{figure}

First, our approach is very computationally intensive both during training and sampling. As is well known, PixelCNN models tend to be very slow to train \citep{oord2016pixel} which can limit the applicability of our method to large scale images. Further, most deterministic inpainting algorithms require a single forward pass of a neural net, while our model (since it is based on PixelCNNs) requires as many forward passes as there are pixels in the image.

Second, our model also has failure modes where it generates implausible inpaintings or inpaintings that do not match the surrounding pixels. A few failure examples are shown in Fig. \ref{failure-examples}.

\section{Conclusion}

In order to address the uncertainty of image inpainting, we have introduced Pixel Constrained CNN, a model that performs probabilistic semantic inpainting by sampling images from a distribution conditioned on the visible pixels. Experiments show that our model generates plausible and diverse completions for a wide variety of regular and irregular masks on the MNIST and CelebA datasets. Further, our model also calculates the likelihood of the inpaintings which, for MNIST, correlates well with the realism of the image completion. 

In future work, it would be interesting to scale our approach to larger images by combining it with methods that aim to accelerate the training and generation of PixelCNN models \citep{ramachandran2017fast, kolesnikov2016pixelcnn, reed2017parallel, menick2018generating}. Further, it would be interesting to explore more sophisticated ways of incorporating the conditional information, such as using attention on the prior and conditional logits or deeply embedding the conditional logits in the model.

\subsubsection*{Acknowledgments}
The authors would like to thank Erik Burton, Jos\'{e} Celaya, Vishakh Hegde and Jin Xu for feedback and helpful discussions. We would also like to thank the anonymous reviewers for helpful suggestions and comments that helped improve the paper. Finally, we would like to thank all the users who took part in our survey.

\bibliography{biblio}

\begin{thebibliography}{}

\bibitem[Afonso et~al., 2011]{afonso2011augmented}
Afonso, M.~V., Bioucas-Dias, J.~M., and Figueiredo, M.~A. (2011).
\newblock An augmented lagrangian approach to the constrained optimization
  formulation of imaging inverse problems.
\newblock {\em IEEE Transactions on Image Processing}, 20(3):681--695.

\bibitem[Arjovsky et~al., 2017]{arjovsky2017wasserstein}
Arjovsky, M., Chintala, S., and Bottou, L. (2017).
\newblock Wasserstein gan.
\newblock {\em arXiv preprint arXiv:1701.07875}.

\bibitem[Ballester et~al., 2001]{ballester2001filling}
Ballester, C., Bertalmio, M., Caselles, V., Sapiro, G., and Verdera, J. (2001).
\newblock Filling-in by joint interpolation of vector fields and gray levels.
\newblock {\em IEEE transactions on image processing}, 10(8):1200--1211.

\bibitem[Barnes et~al., 2009]{barnes2009patchmatch}
Barnes, C., Shechtman, E., Finkelstein, A., and Goldman, D.~B. (2009).
\newblock Patchmatch: A randomized correspondence algorithm for structural
  image editing.
\newblock {\em ACM Transactions on Graphics (ToG)}, 28(3):24.

\bibitem[Bellemare et~al., 2017]{bellemare2017cramer}
Bellemare, M.~G., Danihelka, I., Dabney, W., Mohamed, S., Lakshminarayanan, B.,
  Hoyer, S., and Munos, R. (2017).
\newblock The cramer distance as a solution to biased wasserstein gradients.
\newblock {\em arXiv preprint arXiv:1705.10743}.

\bibitem[Bertalmio et~al., 2000]{bertalmio2000image}
Bertalmio, M., Sapiro, G., Caselles, V., and Ballester, C. (2000).
\newblock Image inpainting.
\newblock In {\em Proceedings of the 27th annual conference on Computer
  graphics and interactive techniques}, pages 417--424. ACM
  Press/Addison-Wesley Publishing Co.

\bibitem[Dahl et~al., 2017]{dahl2017pixel}
Dahl, R., Norouzi, M., and Shlens, J. (2017).
\newblock Pixel recursive super resolution.

\bibitem[Dupont et~al., 2018]{dupont2018generating}
Dupont, E., Zhang, T., Tilke, P., Liang, L., and Bailey, W. (2018).
\newblock Generating realistic geology conditioned on physical measurements
  with generative adversarial networks.
\newblock {\em arXiv preprint arXiv:1802.03065}.

\bibitem[Efros and Freeman, 2001]{efros2001image}
Efros, A.~A. and Freeman, W.~T. (2001).
\newblock Image quilting for texture synthesis and transfer.
\newblock In {\em Proceedings of the 28th annual conference on Computer
  graphics and interactive techniques}, pages 341--346. ACM.

\bibitem[Garnelo et~al., 2018]{garnelo2018neural}
Garnelo, M., Schwarz, J., Rosenbaum, D., Viola, F., Rezende, D.~J., Eslami, S.,
  and Teh, Y.~W. (2018).
\newblock Neural processes.
\newblock {\em arXiv preprint arXiv:1807.01622}.

\bibitem[Goodfellow et~al., 2014]{goodfellow2014generative}
Goodfellow, I., Pouget-Abadie, J., Mirza, M., Xu, B., Warde-Farley, D., Ozair,
  S., Courville, A., and Bengio, Y. (2014).
\newblock Generative adversarial nets.
\newblock In {\em Advances in neural information processing systems}, pages
  2672--2680.

\bibitem[He et~al., 2016]{he2016deep}
He, K., Zhang, X., Ren, S., and Sun, J. (2016).
\newblock Deep residual learning for image recognition.
\newblock In {\em Proceedings of the IEEE conference on computer vision and
  pattern recognition}, pages 770--778.

\bibitem[Iizuka et~al., 2017]{iizuka2017globally}
Iizuka, S., Simo-Serra, E., and Ishikawa, H. (2017).
\newblock Globally and locally consistent image completion.
\newblock {\em ACM Transactions on Graphics (TOG)}, 36(4):107.

\bibitem[Kim et~al., 2019]{kim2019attentive}
Kim, H., Mnih, A., Schwarz, J., Garnelo, M., Eslami, A., Rosenbaum, D.,
  Vinyals, O., and Teh, Y.~W. (2019).
\newblock Attentive neural processes.
\newblock {\em arXiv preprint arXiv:1901.05761}.

\bibitem[Kolesnikov and Lampert, 2016]{kolesnikov2016pixelcnn}
Kolesnikov, A. and Lampert, C.~H. (2016).
\newblock Pixelcnn models with auxiliary variables for natural image modeling.
\newblock {\em arXiv preprint arXiv:1612.08185}.

\bibitem[Kwatra et~al., 2005]{kwatra2005texture}
Kwatra, V., Essa, I., Bobick, A., and Kwatra, N. (2005).
\newblock Texture optimization for example-based synthesis.
\newblock In {\em ACM Transactions on Graphics (ToG)}, volume~24, pages
  795--802. ACM.

\bibitem[Li et~al., 2017]{li2017context}
Li, H., Li, G., Lin, L., and Yu, Y. (2017).
\newblock Context-aware semantic inpainting.
\newblock {\em arXiv preprint arXiv:1712.07778}.

\bibitem[Liu et~al., 2018]{liu2018image}
Liu, G., Reda, F.~A., Shih, K.~J., Wang, T.-C., Tao, A., and Catanzaro, B.
  (2018).
\newblock Image inpainting for irregular holes using partial convolutions.
\newblock {\em arXiv preprint arXiv:1804.07723}.

\bibitem[Liu et~al., 2015]{liu2015faceattributes}
Liu, Z., Luo, P., Wang, X., and Tang, X. (2015).
\newblock Deep learning face attributes in the wild.
\newblock In {\em Proceedings of International Conference on Computer Vision
  (ICCV)}.

\bibitem[Menick and Kalchbrenner, 2018]{menick2018generating}
Menick, J. and Kalchbrenner, N. (2018).
\newblock Generating high fidelity images with subscale pixel networks and
  multidimensional upscaling.
\newblock {\em arXiv preprint arXiv:1812.01608}.

\bibitem[Oord et~al., 2016]{oord2016pixel}
Oord, A. v.~d., Kalchbrenner, N., and Kavukcuoglu, K. (2016).
\newblock Pixel recurrent neural networks.
\newblock {\em arXiv preprint arXiv:1601.06759}.

\bibitem[Pathak et~al., 2016]{pathak2016context}
Pathak, D., Krahenbuhl, P., Donahue, J., Darrell, T., and Efros, A.~A. (2016).
\newblock Context encoders: Feature learning by inpainting.
\newblock In {\em Proceedings of the IEEE Conference on Computer Vision and
  Pattern Recognition}, pages 2536--2544.

\bibitem[Ramachandran et~al., 2017]{ramachandran2017fast}
Ramachandran, P., Paine, T.~L., Khorrami, P., Babaeizadeh, M., Chang, S.,
  Zhang, Y., Hasegawa-Johnson, M.~A., Campbell, R.~H., and Huang, T.~S. (2017).
\newblock Fast generation for convolutional autoregressive models.
\newblock {\em arXiv preprint arXiv:1704.06001}.

\bibitem[Reed et~al., 2017]{reed2017parallel}
Reed, S., Oord, A. v.~d., Kalchbrenner, N., Colmenarejo, S.~G., Wang, Z.,
  Belov, D., and de~Freitas, N. (2017).
\newblock Parallel multiscale autoregressive density estimation.
\newblock {\em arXiv preprint arXiv:1703.03664}.

\bibitem[Shen and Chan, 2002]{shen2002mathematical}
Shen, J. and Chan, T.~F. (2002).
\newblock Mathematical models for local nontexture inpaintings.
\newblock {\em SIAM Journal on Applied Mathematics}, 62(3):1019--1043.

\bibitem[Song et~al., 2017]{song2017image}
Song, Y., Yang, C., Lin, Z., Li, H., Huang, Q., and Kuo, C.-C.~J. (2017).
\newblock Image inpainting using multi-scale feature image translation.
\newblock {\em arXiv preprint arXiv:1711.08590}.

\bibitem[Telea, 2004]{telea2004image}
Telea, A. (2004).
\newblock An image inpainting technique based on the fast marching method.
\newblock {\em Journal of graphics tools}, 9(1):23--34.

\bibitem[van~den Oord et~al., 2016]{van2016conditional}
van~den Oord, A., Kalchbrenner, N., Espeholt, L., Vinyals, O., Graves, A.,
  et~al. (2016).
\newblock Conditional image generation with pixelcnn decoders.
\newblock In {\em Advances in Neural Information Processing Systems}, pages
  4790--4798.

\bibitem[Xu and Teh, 2018]{xu2018controllable}
Xu, J. and Teh, Y.~W. (2018).
\newblock Controllable semantic image inpainting.
\newblock {\em arXiv preprint arXiv:1806.05953}.

\bibitem[Yang et~al., 2017]{yang2017high}
Yang, C., Lu, X., Lin, Z., Shechtman, E., Wang, O., and Li, H. (2017).
\newblock High-resolution image inpainting using multi-scale neural patch
  synthesis.

\bibitem[Yeh et~al., 2016]{yeh2016semantic}
Yeh, R., Chen, C., Lim, T.~Y., Hasegawa-Johnson, M., and Do, M.~N. (2016).
\newblock Semantic image inpainting with perceptual and contextual losses.
\newblock {\em arXiv preprint arXiv:1607.07539}.

\bibitem[Yu et~al., 2018a]{yu2018free}
Yu, J., Lin, Z., Yang, J., Shen, X., Lu, X., and Huang, T.~S. (2018a).
\newblock Free-form image inpainting with gated convolution.
\newblock {\em arXiv preprint arXiv:1806.03589}.

\bibitem[Yu et~al., 2018b]{yu2018generative}
Yu, J., Lin, Z., Yang, J., Shen, X., Lu, X., and Huang, T.~S. (2018b).
\newblock Generative image inpainting with contextual attention.

\bibitem[Zhang et~al., 2016]{zhang2016colorful}
Zhang, R., Isola, P., and Efros, A.~A. (2016).
\newblock Colorful image colorization.
\newblock In {\em European Conference on Computer Vision}, pages 649--666.
  Springer.

\end{thebibliography}

\newpage
\newpage

\appendix

\section{Model Architecture}

The code to reproduce all experiments and results in this paper (including weights of trained models) is available at \url{https://github.com/Schlumberger/pixel-constrained-cnn-pytorch}.

\subsection{MNIST}
\begin{table}[h]
\begin{center}
\begin{tabular}{c c}
\textbf{Prior Network}
\\ \hline \\
Restricted Gated Conv Block, 32 filters, $5 \times 5$ \\
$14\times$ Gated Conv Block, 32 filters, $5 \times 5$ \\
Conv, 2 filters, $1 \times 1$ \\
\end{tabular}
\end{center}

\begin{center}
\begin{tabular}{c c}
\textbf{Conditioning Network}
\\ \hline \\
$15\times$ Residual Blocks, 32 filters, $5 \times 5$ \\
Conv, 2 filters, $1 \times 1$ \\
\end{tabular}
\end{center}
\end{table}

\subsection{CelebA}
\begin{table}[h]
\begin{center}
\begin{tabular}{c c}
\textbf{Prior Network}
\\ \hline \\
Restricted Gated Conv Block, 66 filters, $5 \times 5$   \\
$16\times$ Gated Conv Block, 66 filters, $5 \times 5$   \\
Conv, 1023 filters, $1 \times 1$, ReLU\\
Conv, 96 filters, $1 \times 1$\\
\end{tabular}
\end{center}

\begin{center}
\begin{tabular}{c c}
\textbf{Conditioning Network}
\\ \hline \\
$17\times$ Residual Blocks, 66 filters, $5 \times 5$ \\
Conv, 96 filters, $1 \times 1$ \\
\end{tabular}
\end{center}
\end{table}

\section{Model Training}

\textbf{MNIST}
\begin{itemize}
  \item Optimizer: Adam
  \item Learning rate: 4e-4
  \item $\alpha$: 1
  \item Epochs: 50
\end{itemize}

\textbf{CelebA}
\begin{itemize}
  \item Optimizer: Adam
  \item Learning rate: 4e-4
  \item $\alpha$: 1
  \item Epochs: 60
\end{itemize}

\section{Data}
\subsection{MNIST}
We binarize the MNIST images by setting pixel intensities greater than 0.5 to 1 and others to 0.

\subsection{CelebA}
We quantize the CelebA images from the full 8 bits to 5 bits (i.e. 32 colors). The original (218, 178) images are cropped to (89, 89) and then resized to (32, 32).

\section{Mask Generation Algorithm}

The parameters used for generating masks are $\texttt{max\textunderscore num\textunderscore blobs=4}$, $\texttt{iter\textunderscore min}=2$, $\texttt{iter\textunderscore max}=7$ for both MNIST and CelebA.

\begin{algorithm}[h]
\caption{Generate masks of random blobs.}
\begin{algorithmic}
\REQUIRE Mask height $\texttt{h}$ and width $\texttt{w}$
\REQUIRE \texttt{max\textunderscore num\textunderscore blobs}: maximum number of blobs 
\REQUIRE \texttt{iter\textunderscore min}: min \# of iters to expand blob 
\REQUIRE \texttt{iter\textunderscore max}: max \# of iters to expand blob 
\STATE $\texttt{mask} \gets$ zero array of size $(\texttt{h},\texttt{w})$
\STATE $\texttt{num\textunderscore blobs} \sim \text{Unif}(1, \texttt{max\textunderscore num\textunderscore blobs})$
\FOR{$i = \texttt{1:num\textunderscore blobs}$} 
\STATE $\texttt{num\textunderscore iters} \sim \text{Unif}(\texttt{iter\textunderscore min}, \texttt{iter\textunderscore max})$
\STATE $x_0 \sim \text{Unif}(1, \texttt{w})$
\STATE $y_0 \sim \text{Unif}(1, \texttt{h})$
\STATE $\texttt{mask}[x_0, y_0] \gets 1$
\STATE $\texttt{start\textunderscore positions} \gets [(x_0, y_0)]$
\FOR{$j = \texttt{1:num\textunderscore iters}$} 
\STATE $\texttt{next\textunderscore start\textunderscore positions} \gets []$
\FOR{$\texttt{pos} \ \text{in} \ \texttt{start\textunderscore positions}$} 
\FOR{$x, y \ \text{in} \ \texttt{neighbors}(\texttt{pos})$}
\STATE $p \sim \text{Unif}(0, 1)$
\IF{$p > 0.5$}
\STATE $\texttt{mask}[x, y] \gets 1$
\STATE $\texttt{next\textunderscore start\textunderscore positions} \ \text{append} \ (x, y)$
\ENDIF
\ENDFOR
\ENDFOR
\STATE $\texttt{start\textunderscore positions} \gets \texttt{next\textunderscore start\textunderscore positions}$
\ENDFOR
\ENDFOR
\RETURN $\texttt{mask}$
\end{algorithmic}
\end{algorithm}

\section{More Samples}

\begin{figure}[h]
\begin{center}
\includegraphics[width=1.0\linewidth]{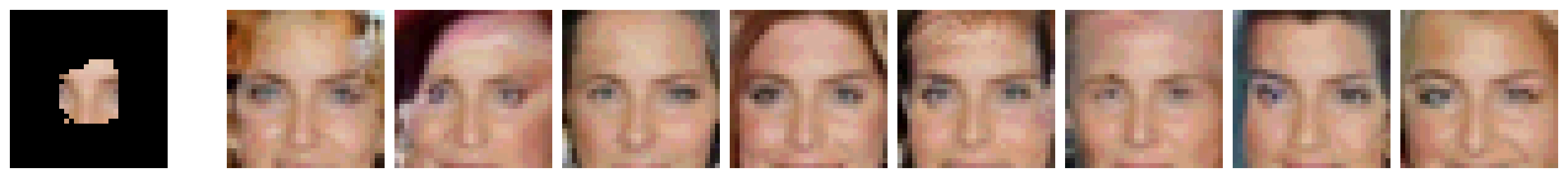}
\includegraphics[width=1.0\linewidth]{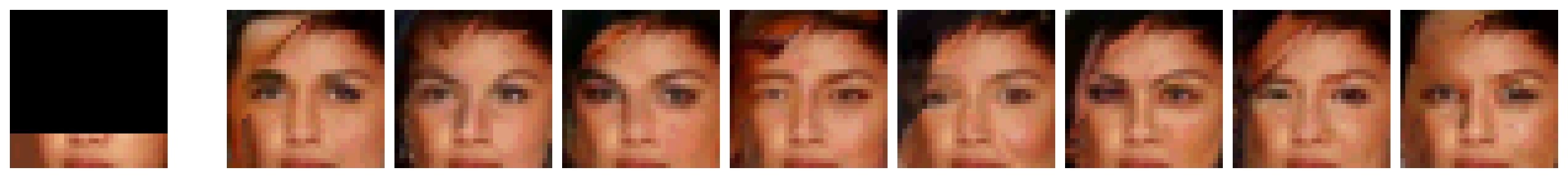}
\includegraphics[width=1.0\linewidth]{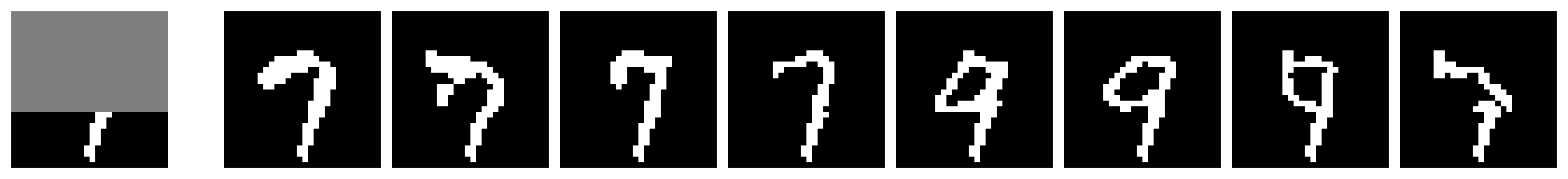}
\includegraphics[width=1.0\linewidth]{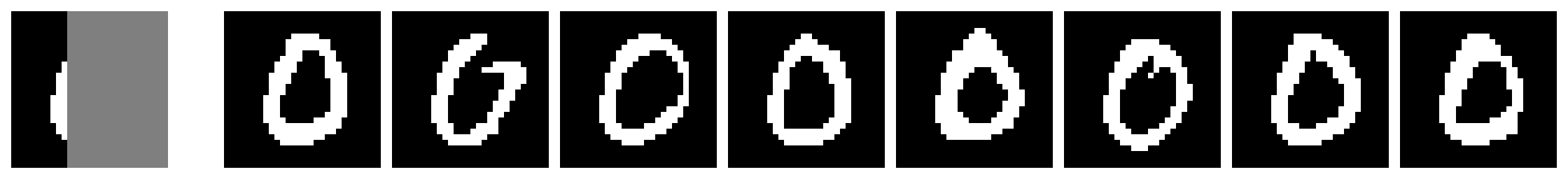}
\includegraphics[width=1.0\linewidth]{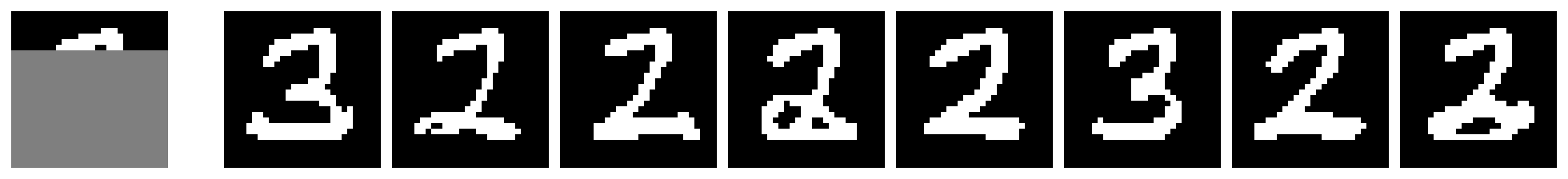}
\end{center}
\caption{Inpaintings.}
\end{figure}

\section{User Study}

The user study was designed to verify if likelihood estimates from the model correlate well with human perception of plausible inpaintings. We include the details of how this test was setup here:
\begin{enumerate}
    \item We randomly selected 100 images and generated 8 inpaintings for each of them.
    \item The samples with highest and lowest likelihood were selected for each of the 100 images.
    \item We then calculated the percentage difference between the highest and lowest likelihood sample for each of the 100 pairs and selected the 15 pairs with the largest difference. This was done to ensure the model assigned significantly different likelihoods to each image.
    \item The user was then presented with the occluded image and was asked to choose which of the generated images they felt was most plausible. They were also given the option to choose neither as some completions were equally good or equally bad.
\end{enumerate}

\begin{figure}[h]
\begin{center}
\includegraphics[width=0.49\linewidth]{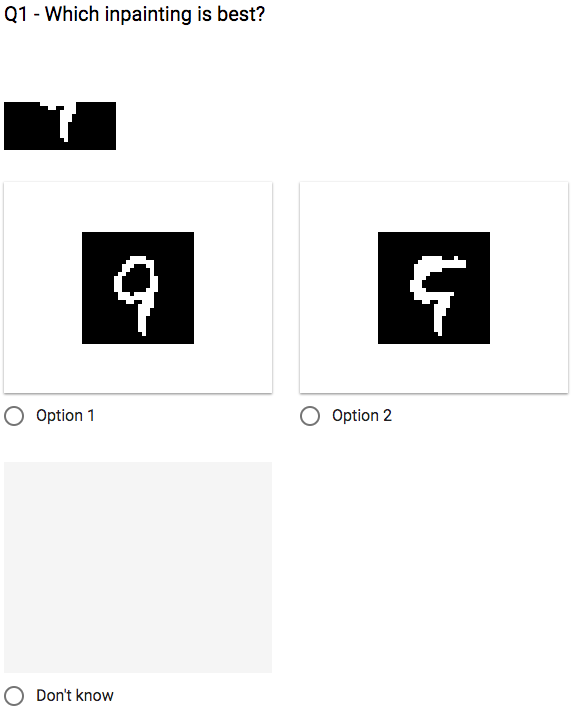}
\includegraphics[width=0.49\linewidth]{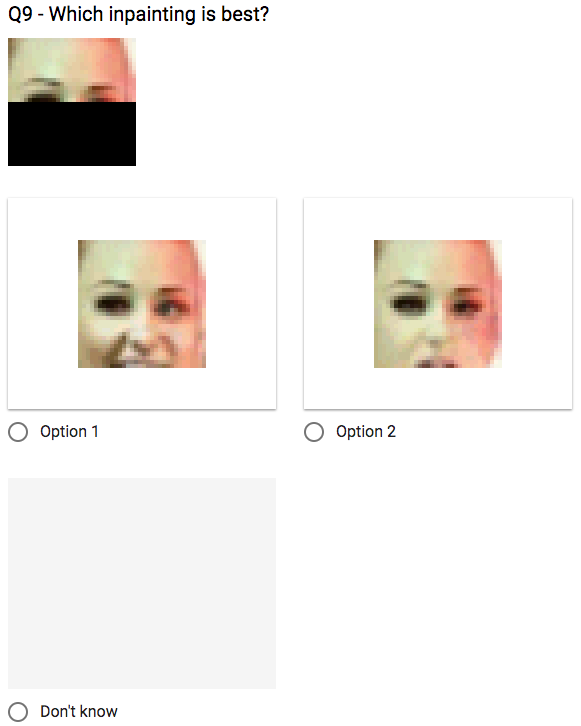}
\end{center}
\caption{Human survey user interface.} \label{user-study}
\end{figure}

Examples of the user interface for the survey are shown in Fig. \ref{user-study}.

\end{document}